\documentclass{article}

\usepackage{microtype}
\usepackage{graphicx}
\usepackage{booktabs} 
\usepackage{dsfont}
\usepackage{makecell}

\usepackage{subcaption}

\usepackage{hyperref}

\usepackage[accepted]{icml2024}

\usepackage{amsmath}
\usepackage{amssymb}
\usepackage{mathtools}
\usepackage{amsthm}
\usepackage{tcolorbox}
\usepackage{xcolor}

\usepackage[capitalize,noabbrev]{cleveref}

\theoremstyle{plain}
\newtheorem{theorem}{Theorem}
\newtheorem{proposition}{Proposition}

\theoremstyle{definition}

\newtheorem{assumption}[theorem]{Assumption}
\theoremstyle{remark}

\DeclareMathOperator*{\argmin}{arg\,min}

\newcommand{\train}{{\mathrm{train}}}
\newcommand{\main}{{\mathrm{main}}}

\DeclareMathOperator{\diag}{\mathbf{diag}}

\newcommand{\aux}{\mathrm{aux}}
\newcommand{\lmain}{L_{\main}}
\newcommand{\laux}{L_{\aux}}
\newcommand{\m}{{\mathrm{m}}}

\usepackage[textsize=tiny]{todonotes}

\usepackage[showdeletions]{color-edits}		
\addauthor[Yu-Guan]{YGH}{blue!80}

\begin{document}

\twocolumn[
\icmltitle{
Careful with that Scalpel: Improving Gradient Surgery with an EMA
}





\begin{icmlauthorlist}
\icmlauthor{Yu-Guan Hsieh}{apple}
\icmlauthor{James Thornton}{apple}
\icmlauthor{Eugene Ndiaye}{apple}
\icmlauthor{Michal Klein}{apple}
\icmlauthor{Marco Cuturi}{apple}
\icmlauthor{Pierre Ablin}{apple}
\end{icmlauthorlist}

\icmlaffiliation{apple}{Apple}

\icmlcorrespondingauthor{Pierre Ablin}{p\_ablin@apple.com}

\icmlkeywords{Machine Learning, ICML}

\vskip 0.3in
]



\printAffiliationsAndNotice{} 

\begin{abstract}
Beyond minimizing a single training loss, many deep learning estimation pipelines rely on an auxiliary objective to quantify and encourage desirable properties of the model (e.g. performance on another dataset, robustness, agreement with a prior).
Although the simplest approach to incorporating an auxiliary loss is to sum it with the training loss as a regularizer, recent works have shown that one can improve performance by blending the gradients beyond a simple sum; this is known as \textit{gradient surgery}.
We cast the problem as a constrained minimization problem where the auxiliary objective is minimized among the set of minimizers of the training loss. 
To solve this bilevel problem, we follow a parameter update direction that combines the training loss gradient and the orthogonal projection of the auxiliary gradient to the training gradient.
In a setting where gradients come from mini-batches, we explain how, using a moving average of the training loss gradients, we can carefully maintain this critical orthogonality property. 
We demonstrate that our method, Bloop, can lead to much better performances on NLP and vision experiments than other gradient surgery methods without EMA.
\end{abstract}

\section{Introduction}
\label{sec:intro}
Overparameterized neural networks trained on large datasets admit multiple solutions with the same optimal training loss \citep{cooper2018loss,li2018visualizing}.
Although these parameters may seem equivalent when viewed through their training loss, they result in different functions, which may exhibit starkly different behaviors on unseen data points.
Practitioners are usually interested in generalization --- one would rather use the network with lower test loss between two networks --- but there are countless other metrics of interest, such as performance on another dataset, robustness, or model calibration.
%
In all of these cases, one aims to train the neural network by minimizing a training loss $\lmain$ while keeping an eye on an auxiliary metric or loss $\laux$.

\textbf{Optimization trade-offs}. Our focus in this paper is on methods that achieve the best possible trade-off between training and auxiliary losses, using a hyper-parameter $\lambda\geq0$ to control that trade-off:
$\lambda=0$ corresponds to training on $\lmain$ exclusively, while increasing $\lambda$ usually decreases $\laux$ at the expense of $\lmain$.
Using the auxiliary loss as a \emph{regularizer} results in the \textit{mixed training method}, arguably the simplest approach to control that trade-off:
\begin{equation}
    \label{eq:regularization}
    \min_{\theta} \lmain(\theta) + \lambda\laux(\theta).
\end{equation}
%
Mixed training, however, runs into optimization issues 
if the directions of the largest curvature of the training loss and that of the auxiliary loss are not aligned --- see \autoref{subsec:conditioning} for an example.

\textbf{The Simple Bilevel Approach. } Provided that modern deep neural networks are inherently overparameterized, leading to multiple minimizers, an ideal solution would be to find the minimizer of $\lmain$ that achieves the smallest auxiliary loss. This corresponds to solving \autoref{eq:regularization} in the limit where $\lambda \to 0$, and can also be expressed as the following \emph{simple bilevel} problem \citep{dempe2010optimality}:
\begin{align}
\label{eq:simple_bilevel}
    \begin{split}
        \min \laux(\theta) ~\text{ s.t. }~\theta\in\arg\min \lmain(\theta).
    \end{split}
\end{align}
Problem~\eqref{eq:simple_bilevel} is a constrained optimization problem on the set of minimizers of $\lmain$, a high-dimensional set with no clear structure, except when $\lmain$ is convex, in which case several provably convergent approaches have been proposed~\citep{sabach2017first,gong2021bi,cao2023projection}. However, to the best of our knowledge, these methods have not been applied to training neural networks, where these convergence guarantees do not hold.

\textbf{Connections to Multi-Task Learning. } The problem of simultaneously optimizing the main and auxiliary loss is also a special case of \emph{multi-task} learning~\citep{caruana1997multitask} involving only two tasks.
Many of the approaches proposed to tackle this problem more efficiently rely on the idea of \emph{gradient surgery}, which stitches together and possibly modify the gradients of both losses when they disagree~\citep{yu2020gradient}.
While multi-task methods tend to treat the two losses equally, we are interested in our work in cases where there is a clear hierarchy between the two.


\vspace{0.2em}
\textbf{Two types of auxiliary losses.}
Auxiliary objectives largely fall into two categories.
The first consists of objectives that guide optimization of the main loss but are not intrinsically meaningful; also known as inductive biases, they are only useful to reach a lower test loss. Weight decay, $\laux = \frac12\|\cdot\|^2$, fits this description: using it improves generalization, but practitioners rarely care about the final norm of their parameters.
The second category of auxiliary losses quantify instead a desirable property: Trading off an increase in the main loss for a decrease in the auxiliary loss might be relevant to applications.
%
%
For instance, the main objective might a loss on a large dataset, whereas the auxiliary objective may be a loss on a smaller, specialized dataset. Ideally, one wishes to achieve a model with high accuracy on both, and hope that the auxiliary loss might also help generalization on the large training set, but both objectives remain meaningful on their own.
Another example is in training neural networks that are also smooth, i.e., with a small Lipschitz constant. This is beneficial for the networks' robustness~\citep{cisse2017parseval}. 
To enforce this during training, one can use a proxy for the Lipschitz constant of the neural network as an auxiliary loss~\citep{tsuzuku2018lipschitz,terjek2019adversarial}.
\looseness=-1

\textbf{Contributions.}
To handle the optimization tradeoff between main and auxiliary losses, we introduce in \autoref{sec:algo} the
\textbf{Bloop} (\textbf{B}i\textbf{L}evel \textbf{O}ptimization with \textbf{O}rthogonal \textbf{P}rojection) method.
Our method is inspired by the simple bilevel problem, but similar to the regularization approach, has a tunable hyperparameter, $\lambda$, to control the trade-off between losses.
At the heart of the method is a projection of the auxiliary gradient to be orthogonal to the primary loss gradient.
We first provide a theoretical justification for this approach in the full-batch case. In the stochastic setting, we rely on an exponential moving average (EMA) of the training gradient to estimate the projection direction, and retain most of the full-batch theoretical properties.\looseness=-1
In \autoref{sec:theory}, we analyze Bloop's stationary points, and show that they are first-order stationary points of the simple bilevel problem.
We demonstrate the convergence of the iterates towards the stationary points of the training loss, under appropriate hypothesis on the step size and the EMA accumulation factor, highlighting the importance of the EMA.
%
In \autoref{sec:related}, we discuss related methods that perform variants of gradient surgery.
In \autoref{sec:expe}, we explore the applicability of our method to a variety of tasks: training network parameters with an explicit bias; multi-task learning; training language models to perform well on a large generic dataset and a small specific dataset. In our experiments, Bloop exhibits a better Pareto front than both the mixed method and multi-task methods that do not use an EMA.
\section{The Bloop Algorithm}

In this section, we introduce Bloop, a simple and intuitive iterative algorithm to optimize two losses simultaneously. We then discuss how the method can be extended to address stochasticity in the gradients, and multi-level optimization.

\label{sec:algo}
\subsection{Full-batch setting and main intuition}
At each step, Bloop builds a parameter update direction $d\in \mathbb{R}^p$ which is then fed to an optimizer (e.g. Adam~\citep{kingma2014adam}) in order to converge to the solution of \autoref{eq:simple_bilevel}.
For instance, the gradient descent optimizer would iterate $\theta \leftarrow \theta -\eta d$.
At the current iterate $\theta$, we let $g_{\main} = \nabla \lmain(\theta)$ and $g_{\aux} = \nabla \laux(\theta)$.

We design our direction from first principles.
We seek a direction in the span of these two gradients, $d = \omega g_\main + \lambda g_\aux$ with $\omega$ and $\lambda$ two scalars.
Our primary goal is to make progress on the main loss at the same speed as gradient descent; hence we target $\lmain(\theta - \eta d) \simeq \lmain(\theta - \eta g_\main).$

At the first order in the step-size $\eta$, we see that the component of the direction in the direction $g_\main$ should be the same as that of $g_\main$, i.e., we want $\langle d, g_\main\rangle = \|g_\main\|^2$.
This gives the equation $(1 - \omega) \|g_\main\|^2 = \lambda \langle g_\main, g_\aux\rangle$.
Our secondary goal is the optimization of the auxiliary loss, hence we impose that the coefficient in front of $g_\aux$ is positive, i.e. that $\lambda >0$.
These two conditions alone give us our update rule: we find that such a direction is necessarily
\begin{tcolorbox}[colback=white, colframe=black, boxrule=1pt, arc=5mm, boxsep=1mm, left=0mm, right=2mm, valign=center]
\vspace{-1.5em}
\begin{align}
    \label{eq:direction}
    \begin{split}
            d = g_\main + \lambda \pi(g_\aux, g_\main), \text{  where}~~~\\
             \pi(g_\aux; g_\main) = g_\aux - \frac{\langle g_\aux, g_\main \rangle}{\|g_\main\|^2}g_\main
    \end{split}
\end{align}
\end{tcolorbox}

\begin{figure}[t]
    \centering
\includegraphics[width=.75\columnwidth]{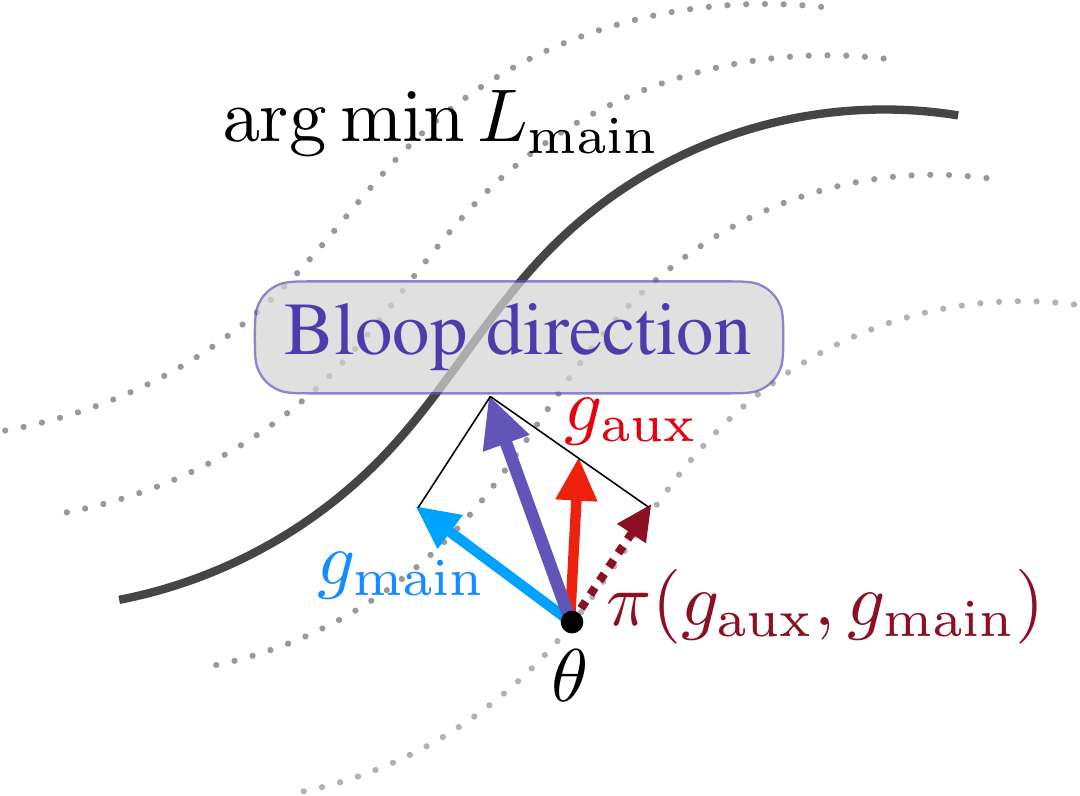}
    \caption{\textbf{Principle of the Bloop method}: the direction we follow is the sum of the gradient of the main loss $g_\main$, and of the projection of the gradient of the auxiliary loss, orthogonal to $g_\main$.
    This enforces that, at the first order, following this direction yields the same decrease in $\lmain$ as following $g_\main$.}
    \label{fig:simple_bilevel_principle}
\end{figure}

Hyperparameter $\lambda\geq0$ trades-off the two objectives, and $\pi(g_\aux; g_\main)$ is the projection of $g_\aux$ orthogonal to $g_\main$.
This direction admits an intuitive explanation: since we primarily want to optimize the main loss, we follow $g_\main$; 
the projection part is aligned with $g_{\aux}$, and does not interfere with $g_\main$ thanks to the orthogonality condition.
Moreover, the fact that $\langle d, g_\main\rangle = \|g_\main\|^2$ means that following this direction does not change the optimization with respect to $\lmain$ when step-sizes are small. Specifically, we write down the Taylor expansion at the first order
\begin{align}
    \label{eq:descent_gd}
    \begin{split}
    \lmain(\theta - \eta d) &\simeq\lmain(\theta) - \eta \langle g_\main, d\rangle, \\
   \text{\color{gray}(Orthogonality)\hspace{0.5em}} &\simeq\lmain(\theta) - \eta \|g_\main\|^2.
    \end{split}
\end{align}
This is the same as standard gradient descent where $d=g_\main$.
\autoref{fig:simple_bilevel_principle} illustrates the geometric principle of Bloop.

\subsection{Stochastic extension for large-scale problems}
When dealing with neural networks trained over large datasets, the losses are written as sums over many samples:
\begin{equation*}
    \lmain(\theta) = \frac1n\sum_{i=1}^n\lmain^i(\theta),\enspace \laux(\theta) = \frac1m\sum_{j=1}^m \laux^j(\theta).
\end{equation*}
In practice, we can only use a mini-batch of gradients to make progress on the problem, as the computation of the full-batch gradient of these losses is out of the question.
Concretely, we assume that we have computed the two mini-batch gradients $g^{\mathrm{batch}}_\main$, $g^{\mathrm{batch}}_\aux$, which are by design unbiased estimators of the full-batch gradients:
\begin{equation*}
    \mathbb{E}[g^{\mathrm{batch}}_\main] = g_\main ~\text{  and  }~\mathbb{E}[g^{\mathrm{batch}}_\aux] = g_\aux.
\end{equation*}
In the above, the expectation is taken over the randomness of the mini-batch choice.
Extending the direction $d$ to this \emph{stochastic} setting is not straightforward, and careful design makes a big difference in the final performance.
A key insight behind standard, single-level, stochastic gradient descent on $\lmain$ is that, for small step sizes, it has on average the same decrease as gradient descent:
\begin{align*}
    \begin{split}
    \mathbb{E}[\lmain(\theta - \eta g^{\mathrm{batch}}_\main)] &\simeq\lmain(\theta) - \eta \mathbb{E}[\langle g_\main, g^{\mathrm{batch}}_\main\rangle] \\
   \text{\color{gray}(Linearity of dot)\hspace{2em}} &\simeq\lmain(\theta) - \eta \langle g_\main, \mathbb{E}[g^{\mathrm{batch}}_\main]\rangle\\
   \text{\color{gray}(Unbiased gradient)\hspace{1em}} &\simeq\lmain(\theta) - \eta\|g_\main\|^2\\
    \end{split}
\end{align*}
We want to preserve this behavior as much as possible.
A first idea is simply to plug the mini-batch gradients in \autoref{eq:direction}, i.e. consider
$$
d^{\mathrm{batch}}_{\mathrm{simple}} = g^{\mathrm{batch}}_\main + \lambda \pi(g^{\mathrm{batch}}_\aux;  g^{\mathrm{batch}}_\main).
$$
\textbf{The pitfall of projecting on stochastic gradients.}
The main issue with the above method is that the projection is nonlinear with respect to its second argument: in general, $\mathbb{E}[\pi(g^{\mathrm{batch}}_\aux; g^{\mathrm{batch}}_\main)] \neq \pi(g_\aux; g_\main)$.
As a consequence, it is not true anymore that $\langle d^{\mathrm{batch}}_{\mathrm{simple}}, g_\main\rangle = \|g_\main\|^2$, even in expectation, which in turn leads to a behavior starkly different from SGD on $\lmain$.
We can improve this intuition using a simplified model of the training dynamics.
Assume that $g^{\mathrm{batch}}_\main = g_\main + \sigma \varepsilon$, where $\varepsilon\sim \mathcal{N}(0, I)$ is the random gradient noise, and $\sigma >0$ is the noise variance.
In the limit where $\sigma$ is large in front of $\|g_\main\|$, we get that on average $\mathbb{E}_{\varepsilon}[\pi(g_\aux^{\mathrm{batch}}; g_\main^{\mathrm{batch}})] = (1 - \frac1p)g_\aux^{\mathrm{batch}}$ with $p$ the parameter's dimension.
Therefore, the simple direction is on average $d^{\mathrm{batch}}_{\mathrm{simple}} =  g^{\mathrm{batch}}_\main + \lambda (1 - \frac1p)g_\aux^{\mathrm{batch}}$. We recover the same direction as that of the mixed training method, with a new $\lambda' = \lambda (1- \frac1p)$, and the orthogonalization becomes useless.\looseness=-1

\begin{figure}[t]
    \centering
\includegraphics[width=.75\columnwidth]{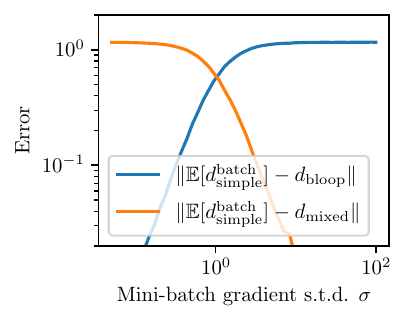}
    \caption{\textbf{Effect of randomness on the projection}: We fix the dimension of the parameter space to $p=100$, and draw both $g_{\text{main}}$ and $g_{\text{aux}}$ from the Gaussian distribution $\mathcal{N}(\mathbf{0}, I)$. 
These two vectors are fixed in the remainder of the experiment.
We draw $g_{\text{main}}^{\text{batch}} \sim g_{\text{main}} + \sigma \mathcal{N}(\mathbf{0}, I)$ and use Monte-Carlo simulation to estimate $\mathbb{E}[d_{\text{simple}}^{\text{batch}}] =g_{\text{main}} + \mathbb{E}[\pi(g_{\text{aux}}; g_{\text{main}}^{\text{batch}})]$.
We compare its value against $d_{\text{bloop}}=g_{\text{main}} + \pi(g_{\text{aux}}; g_{\text{main}})$, its theoretical value when $\sigma=0$ (the target direction), and $d_{\text{mixed}}=g_{\text{main}} + (1-1/100) g_{\text{aux}}$, its theoretical value when $\sigma$ tends to infinity.
We see that the $\mathbb{E}[d^{\text{batch}}_{\text{simple}}]$ becomes closer to the gradient of the mixed method when the noise starts to dominate.}
    \label{fig:noise_effect}
\end{figure}

In order to illustrate this intuition, we conduct a synthetic experiment, explained in \autoref{fig:noise_effect}. 

\textbf{The EMA solution.}\enspace The previous analysis indicates that we need a better estimate of $g_\main$ than the mini-batch gradient.
A simple solution to this is to use an Exponential Moving Average (EMA) of the previous batch gradients, $g_\main^\mathrm{EMA}$, which is updated at each iteration by doing $g_\main^\mathrm{EMA}\leftarrow (1-\rho)g_\main^\mathrm{EMA} + \rho g_\main^{\mathrm{batch}}$, with $\rho\in[0, 1]$ a parameter that controls the speed of the EMA.
This can be a much better estimator of $g_\main$ than $g_\main^{\mathrm{batch}}$, because it averages gradients over the optimization trajectory, drastically reducing the variance.
Intuitively, we need to accumulate the EMA faster than the speed of the optimization algorithm that updates the parameters. Hence, $\rho$ should be greater than the step-size $\eta$.
%
We use this gradient EMA solely in the projection, and propose the direction
\begin{tcolorbox}[colback=white, colframe=black, boxrule=1pt, arc=5mm, boxsep=1mm, left=0mm, right=2mm, valign=center]
\vspace{-1em}
\begin{align}
    \label{eq:stochastic_direction}
    \begin{split}
            d^{\mathrm{batch}} = g^{\mathrm{batch}}_\main + \lambda \pi(g^{\mathrm{batch}}_\aux;g^{\mathrm{EMA}}_\main)
    \end{split}
\end{align}
\end{tcolorbox}
We do not replace the first $g^{\mathrm{batch}}_\main$ in the formula by the EMA, because $d^{\mathrm{batch}}$ is an optimization \emph{direction}, that is then plugged into any optimizer like Adam, which will use a smart adaptive step to reach the solution quickly.
Since the EMA does not depend on the current batch, and the projection is linear with respect to its first argument, we have that $\mathbb{E}[d^{\mathrm{batch}}] = g_\main + \lambda \pi(g_\aux; g^{\mathrm{EMA}}_\main)$, and as a consequence, the expected decrease on $\lmain$ following this direction is 
$
\mathbb{E}[\lmain(\theta - \eta d^{\mathrm{batch}})]\simeq\lmain(\theta) - \eta\|g_\main\|^2 +\eta\lambda \langle \pi(g_\aux; g^{\mathrm{EMA}}_\main), g_\main\rangle
$.
When the EMA accumulation $g^{\mathrm{EMA}}_\main$ is close to $g_\main$, the last term becomes small because the two vectors are approximately orthogonal. Thus,
$$
\mathbb{E}[\lmain(\theta - \eta d^{\mathrm{batch}})] \simeq \lmain(\theta) - \eta\|g_\main\|^2,
$$
and we recover the same behavior as SGD on $\lmain$.
The new direction is no longer in the span of $(g^{\mathrm{batch}}_\main$, $g^{\mathrm{batch}}_\aux)$ because it also has a component in the direction of $g^{\mathrm{EMA}}_\main$.

The theory presented in the next section clearly highlights the importance of this EMA, and in our experiments, we find that this simple EMA modification drastically improves the performance of the algorithm on a variety of tasks.
In fact, we found that in many cases, standard multi-task methods without EMA have very similar performances to the mixed training method.

\autoref{alg:bloop} gives the full pseudo-code of the Bloop method. 
We use \texttt{optax}-like notations~\citep{deepmind2020jax} for the optimizer, which is abstracted as a method that, given a direction $d$, current parameters $\theta$ and a state $s$ containing all its hyper-parameters like learning rate and internal state like EMAs for adaptive methods, returns the updated parameters $\theta$ and updated state $s$. 
\begin{algorithm}[t]
\begin{algorithmic}
   \STATE {\bfseries Input:} Hyperparameter $\lambda$, EMA parameter $\rho$, initial parameters $\theta$, optimizer $\texttt{optim}$, optimizer state $s$, initial EMA $g_\main^{\mathrm{EMA}}$
   \FOR{$t = 0,\dots, T-1$}
   \STATE 
   Sample gradients $g_\main^{\mathrm{batch}}$, $g_\aux^{\mathrm{batch}}$

   Compute the Bloop direction $d^{\mathrm{batch}}$ using \autoref{eq:stochastic_direction}

   Update $\theta, s \leftarrow \texttt{optim}(d^{\mathrm{batch}}, \theta, s)$
   
   Update EMA: $g_\main^{\mathrm{EMA}}\leftarrow (1-\rho)g_\main^{\mathrm{EMA}} + \rho g_\main^{\mathrm{batch}}$
   \ENDFOR
\end{algorithmic}
\caption{The Bloop algorithm}
\label{alg:bloop}
\end{algorithm}


\subsection{Extension to multi-level hierarchical optimization}
Our algorithm can be extended to multi-level optimization, where we have more than two losses and they have a \emph{hierarchy}.
For simplicity, we present here the case with $3$ losses: $\lmain$, $\laux^1$ and $\laux^2$.
The hierarchy means that we minimize $\lmain$, and then, among this set of minimizers, we  minimize $\laux^1$. Finally, we minimize $\laux^2$ among this new set.
This gives the trilevel optimization problem:
\begin{align}
    \begin{split}
        & \min \laux^2(\theta)\text{ s.t. } \\
        \theta &\in \left(\argmin \laux^1(\theta) \text{ s.t. }\theta \in\argmin \lmain(\theta)\right)
    \end{split}
\end{align}
Our algorithm can be straightforwardly extended to this case by following a Gram-Schmidt like orthogonalization process: letting $g_\main$, $g_\aux^1$ and $g_\aux^2$ the gradients of the three losses, we go in the direction 
$$
d = g_\main +\lambda^1 \pi(g_\aux^1; g_\main) + \lambda^2 \pi(g_\aux^2; (g_\main, g_\aux^1))
$$
where $\pi(g_\aux^2; (g_\main, g_\aux^1))$ is the projection of $g_\aux^2$ on the orthogonal of the span of $(g_\main, g_\aux^1)$.
Thanks to orthogonality, this direction satisfies 
$\langle d,  g_\main\rangle = \|g_\main\|^2$; hence in terms of optimization with respect to $\lmain$, the direction behaves just like $g_\main$, and $\langle d, g_\aux^1\rangle = \langle g_\main +\lambda^1 \pi(g_\aux^1; g_\main), g_\aux^1\rangle$; hence in terms of optimization with respect to $\laux^1$, the direction behaves just like the bilevel direction $d$ introduced in \autoref{eq:direction}.

\section{Theoretical Analysis}
\label{sec:theory}
This section aims at understanding the theoretical properties of the proposed direction in the full-batch and the mini-batch settings by linking it with the simple bilevel problem (\autoref{eq:simple_bilevel}).
All the proofs are deferred to \cref{apx:proofs}.

\subsection{Approximate stationary points of Bloop}

At a solution to the simple bilevel problem, we have $\nabla \lmain(\theta) = 0$, hence the solutions to the bilevel problem are also solutions of
$$
\min L_\aux(\theta)\text{ s.t. } \nabla \lmain(\theta) = 0.
$$
The Lagrangian for this equation is 
$
\mathcal{L}(\theta, v) = L_\aux(\theta) - \langle v, \nabla \lmain(\theta) \rangle$
with $v\in\mathbb{R}^p$ the Lagrange multiplier.
Accordingly, the first-order optimality conditions are $g_\main = 0$ and that there exists $v$ such that $g_\aux = \nabla^2L_\main(\theta)v$.
A first natural question to ask is whether the direction that we propose in \autoref{eq:direction} cancels at these points.
However, the projection is ill-defined when $g_{\main} = 0$. We thus assume that $\|g_{\main}\|$ is positive hereinafter and focus on the case where $d$ is small but non-zero.\footnote{Although we can simply set $d=\lambda g_\aux$ when $g_\main=0$, the study of this particular case is straightforward and gives little insight on the general case. We therefore omit it here.}
To analyze this, we introduce the following assumption.  
\begin{assumption}[Local Error Bound~\citealp{luo1993error}]\label{assum:local_error_bound}
    There exists $c>0$ such that for $\varepsilon$ small enough and for any $\theta$ satisfying $\|g_\main(\theta)\|\leq\varepsilon$, we have $$
    \mathrm{Dist} (\theta, \nabla \lmain^{-1}(\{0\}))\leq c\|g_\main\|.$$
\end{assumption}
This local error bound condition is implied by a local Polyak-Lojasiewicz inequality, which is verified, for instance, for overparameterized least-squares and some neural network loss functions~\citep{liu2022loss}.
With this in hand, we are now ready to present our
result regarding the approximate first-order stationary points of the full-batch Bloop method.
\begin{proposition}[Stationary points]
\label{prop:stat}
    If $d$ in \autoref{eq:direction} is such that $\|d\|\leq\varepsilon$, then we have
    $\|g_\main\|\leq \varepsilon$. Moreover if \Cref{assum:local_error_bound} holds, the Hessian of $\lmain$ is $M-$Lipschitz, and $\varepsilon$ is small enough,
    then there exists $v\in\mathbb{R}^p$ such that $$\|g_\aux - \nabla^2L_\main(\theta)v\|\leq(\lambda^{-1} + Mc^2\|g_\aux\|/2)\varepsilon.$$
    Conversely, given a point $\theta^*$ that satisfies the first order optimality conditions of \autoref{eq:simple_bilevel}, we have that $\lim_{\varepsilon\to 0}d(\theta^* + \varepsilon v) = 0$ where $v$ is the Lagrange multiplier.
\end{proposition}
In short, \cref{prop:stat} relates the (approximate) stationary points of Bloop to the (approximate) stationary points of the bilevel problem.
Moreover, as an immediate consequence of the proposition, we see that we additionally assume $\laux$ to be Lipschitz continuous, the limit points of Bloop must be stationary points of the simple bilevel problem.

\subsection{Convergence of stochastic Bloop}
Our main theorem is a convergence result of the \emph{stochastic version} of Bloop.
It clearly highlights the role of the EMA: without EMA, obtaining such results would be impossible.
\begin{theorem}[Convergence of Bloop]
\label{thm:convergence}
Consider the Bloop method in the stochastic setting with the SGD optimizer. Let $\rho$ be the EMA parameter and $\eta$ be the step-size of the algorithm.
Assume that (i) $\lmain$ is $L$-smooth, (ii) the stochastic directions are uniformly bounded, i.e., $\|d^t\|\leq D$ for all $t$, (iii) the variance of the gradients of $\lmain$ is bounded with $\mathbb{E}_i[\|\nabla \lmain^i(\theta) -\nabla \lmain(\theta)\|^2 \leq C^2$, and (iv) the auxiliary gradients are bounded as $\|\nabla\laux(\theta)\|\leq B$.
Then, for a number of iterations $T$, taking a step size $\eta\simeq T^{-\frac34}$ and an EMA parameter $\rho \simeq \eta^{\frac23}$ gives 
$$
\frac1T\sum_{t=0}^{T-1}
\mathbb{E}[\|\nabla \lmain(\theta^t)\|^2] = O(T^{-\frac14})
$$
If $\lmain$ is additionally $\mu$-PL~\citep{karimi2016linear}, we have 
$$
\resizebox{\columnwidth}{!}{$%
\mathbb{E}[\lmain(\theta^T)-\min\lmain]\leq (1 - 2\eta\mu)^T\lmain(\theta^0) + O(\eta^{\frac13}).$}
$$
\end{theorem}
\cref{thm:convergence} demonstrates the convergence of stochastic Bloop either in terms of the expected gradient norm or the expected optimiality gap.
In spirit, this suggests that the Bloop iterate would end up being arbitrarily close to the stationary points of $\lmain$.
The theorem also instructs us on the role of the EMA coefficient $\rho$ compared to the learning rate $\eta$. 
We see that we should take $\rho$ to be slightly larger than $\eta$: in this regime, the gradient EMA $g_\train^{\mathrm{EMA}}$ is a good approximation of $g_{\train}$.

Also note that this result differs significantly from those obtained in the multi-task learning literature, which show convergence of the algorithms to points where either both losses are minimized or where their gradients are opposed~\citep{yu2020gradient}. 
Here, even in the extreme case where losses are the exact opposite ($\laux = -\lmain$), full-batch Bloop provably converges to the minimizers of $\lmain$ under PL condition. This is not a surprise since in that case, the projection $\pi(g_{\aux}, g_\main)$ cancels and the iterates of Bloop are that of gradient descent on $\lmain$.

Unlike \citet{gong2021bi}, we do not demonstrate the convergence of our algorithm to the KKT points of the simple bilevel problem.
Our results are thus weaker in that regard, albeit in a different setting since \citet{gong2021bi} are not in the stochastic case. 

\subsection{Conditioning compared to regularization method}
\label{subsec:conditioning}
We illustrate below that the regularization method can lead to poorly conditioned problems, resulting in hard optimization problems, while our method alleviates this.
For this, we take the following simple $2$D example, where $\theta=(a, b)$:
\begin{equation*}
    \lmain(\theta) = \frac12a^2\,, ~~\laux(\theta) = \frac12((a-1)^2 + b^2).
\end{equation*}
The solution to the bilevel problem is $\theta^* = 0$, while the solution to the regularized problem is $\theta=(\lambda/(1+\lambda), 0)$.
We recover the same solution in the limit $\lambda \to 0$.
However, the Hessian of the regularized problem is $\diag(1 + \lambda, \lambda)$; hence the conditioning of the regularized problem is $1+1/\lambda$ which goes to infinity as $\lambda\to 0$.
In view of this, the regularized method either converges to a point far from the solution ($\lambda$ large) or converges slowly ($\lambda$ small).
On the contrary, the projection method goes in the direction $d=(a, \lambda b)$. This is equivalent to gradient descent on a quadratic loss with the correct $\theta^*$ minimizer --- \emph{regardless of $\lambda$} --- and Hessian equal to $\diag(1, \lambda)$, which is well conditioned when $\lambda$ is not too far from $1$.\looseness=-1

\setlength{\tabcolsep}{4pt}
\begin{table}[t]
\caption{Comparison of similar gradient surgery methods for the 
two tasks setting.
For brevity, we write $g_\m := g_\main$ and $\phi := \cos ( g_\m, g_\aux) = \frac{\langle g_\m, g_\aux \rangle}{\|g_\m\|\|g_\aux\|}$. $\bar{(\cdot)}$ indicates that EMA has been applied, and $\psi$ is a dynamic barrier function described in \citep{gong2021bi}.}
\label{sample-table}
\vskip 0.15in
\begin{center}
{\small

\begin{tabular}{lcccr}
\hline
\abovespace\belowspace
Method & {Modified Direction}  \\
\hline
\abovespace
Bloop (ours)  &  $g_\m + \lambda\left( g_\aux - \frac{\langle g_\aux, \bar{g}_\m\rangle}{\|\bar{g}_\m\|^2}\bar{g}_\m \right)$ \\[0.3cm]  

Mixed (Regularized) &  $g_\m + \lambda g_\aux$ \\[0.3cm] 

\begin{tabular}[c]{@{}l@{}}A-GEM\\ 
{\scriptsize \citet{chaudhry2018efficient}}\end{tabular} &  $g_\m - \frac{\min(0, \langle g_\m, g_\aux \rangle)}{\|g_\aux\|^2}g_\aux$\\[0.3cm]

\begin{tabular}[c]{@{}l@{}}Dynamic Barrier \\ {\scriptsize \citet{gong2021bi}}\end{tabular}  & $g_\aux + \max(0, \frac{\psi(\theta) - \langle g_\m, g_\aux \rangle}{\|g_\m\|^2})g_\m$   \\[0.35cm]


\begin{tabular}[c]{@{}l@{}}MTL-MOO\\ 
{\scriptsize  \citet{sener2018multi}}\end{tabular} &  
\begin{tabular}[c]{@{}l@{}} $\frac{\langle g_\m-g_\aux, g_\aux \rangle}{\|g_\m-g_\aux\|^2} g_\m + $\\ $(1-\frac{\langle g_\m-g_\aux, g_\aux \rangle}{\|g_\m-g_\aux\|^2})g_\aux$
\end{tabular} \\[0.3cm]


\begin{tabular}[c]{@{}l@{}}Cosine Similarity\\ {\scriptsize \citet{du2018adapting}}\end{tabular}    &      $g_\m + g_\aux \max(0, \phi)$   \\[0.3cm]


\begin{tabular}[c]{@{}l@{}}GradVac\\ {\scriptsize \citet{wang2021gradient}}\end{tabular}      &       $g_\m + \frac{\|g_\m\| \left(\bar{\phi}\sqrt{1-\phi^2}-\phi\sqrt{1-\bar{\phi}^2}\right)}{\|g_\aux \sqrt{1-\bar{\phi}^2}\|}$ \\[0.3cm]

\begin{tabular}[c]{@{}l@{}}PCGrad\\ {\scriptsize \citet{yu2020gradient}}\end{tabular}       &        \begin{tabular}[c]{@{}l@{}}$ g_\m -\min(0,\langle g_\aux, g_\m \rangle)\frac{g_\m}{\|g_\m\|^2}$ \\ $+ g_\aux -\min(0,\langle g_\aux, g_\m \rangle)\frac{g_\aux}{\|g_\aux\|^2}$\end{tabular}
\\[0.3cm]

\begin{tabular}[c]{@{}l@{}}Meta-Balance\\ {\scriptsize \citet{he2022metabalance}}\end{tabular}       &  $g_\m + \frac{\|g_\m\|}{\|g_\aux\|}g_\aux$  \\[0.3cm]  
\hline
\end{tabular}
}
\end{center}
\vskip -0.1in
\end{table}

\section{Related Works}
\label{sec:related}

Our work sits at the intersection of two fields of machine learning: the solution of the simple bilevel problem and multi-task learning. There are however a number of differences between the two. In particular, in the multi-task learning problem each task is considered jointly whereas in the bilevel setting there is a hierarchy to the primary and auxiliary objectives. Another key difference is in the notion of task versus auxiliary objective. A task typically requires a dataset as input, whereas an auxiliary objective is more general and can incorporate losses without the need for data, such as the $L^2$ norm in weight decay.

Given the similarity, a number of gradient surgery methods that have been proposed in multi-task literature can be used to minimize both the main and the auxiliary objectives. We summarize the most relevant ones in \cref{sample-table}.
Some works try to leverage the auxiliary loss to obtain improvements on the main loss only~\citep{du2018adapting,dery2021auxiliary}.

The Dynamic Barrier (DB) algorithm of \citet{gong2021bi}, as detailed in \cref{sample-table}, uses a similar orthogonal projection as in our proposal.
It provably solves the bilevel problem.
However, DB includes an additional barrier function, $\phi$ e.g. $\phi = \|g_\aux\|^2$, to control the trade-off between objectives, whereas we use a scalar, $\lambda$, similar to regularization methods, for this purpose. The other main differences between our proposal and the DB method are that we always use the projection, rather than conditioning on $\langle g_\m, g_\aux\rangle$, and most importantly, we use an EMA of main gradients to compute the projection, rather than the stochastic gradient. With $\phi = \|g_\aux\|^2$ and without the conditional update or EMA, the approaches would be the same. \citet{gong2021bi} do not discuss stochastic extensions of the method, which is of key importance to practitioners.

\citet{yu2020gradient} propose PCGrad, which, as shown in \cref{sample-table}, can be regarded as a symmetrized version of our method. Unlike our method, the projection is again conditioned. 
Concretely, the parameters are updated in the direction of the combined gradient $g_\main + g_\aux$ when they are aligned, and projections are performed when this is not the case.
The gradient alignment condition and the symmetry between the gradients implies that the algorithm does not solve the bilevel problem; instead \citep[Thm.1]{yu2020gradient} show that it minimizes the sum of the two losses or finds a point where $g_\aux$ and $g_\main$ go in opposite directions. Similarly to the DB method, no EMA is used in the projection.


\section{Experiments}
\label{sec:expe}

\begin{figure*}[t]
    \centering
    \begin{subfigure}[t]{0.49\textwidth}
    \includegraphics[width=.49\columnwidth]{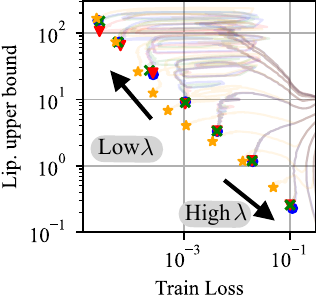}
    \hfill
    \includegraphics[width=.49\columnwidth]{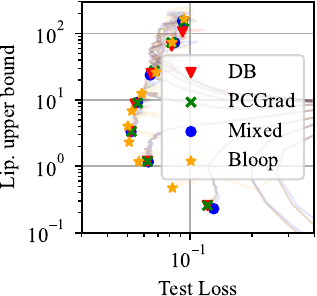}
        \caption{Training an MLP on MNIST with an auxiliary loss that is a proxy for its Lipschitz constant.}
    \label{fig:lipschitz}
    \end{subfigure}
    \hfill
    \begin{subfigure}[t]{0.49\textwidth}
    \includegraphics[width=.49\columnwidth]{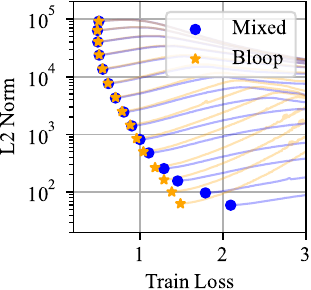}
    \hfill
    \includegraphics[width=.49\columnwidth]{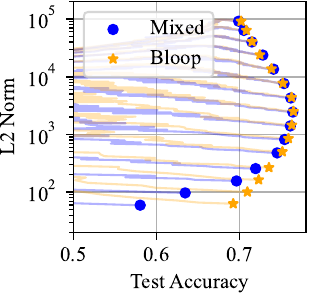}
        \caption{Training a ResNet50 on Imagenet with squared L2 norm as the auxiliary loss.}
    \label{fig:imagenet_l2}
    \end{subfigure}
    \caption{Trade-offs between the main and the auxiliary objectives in problems where the auxiliary loss is used to impose an explicit bias on the neural network.
    The symbols correspond to the parameters reached at the end of training and form a Pareto front, the transparent curves are the training trajectories.
    Bloop achieves a better trade-off than the other methods, which all perform similarly here.
    }
    \label{fig:exp-bias}
\end{figure*}

In this section we demonstrate the effectiveness of Bloop via numerical experiments on problems of three distinct categories: the use of auxiliary loss for imposing an explicit bias, multi-task learning, and joint dataset training.
For each of these experiments, we use an optimizer with hyperparameters that work well for the minimization of solely the main loss, and never change these hyperparameters.
As for the EMA parameter of Bloop, we take it as $\rho = 0.01$ in all experiments unless otherwise stated.
Further experimental details can be found in \cref{apx:exp}.

Note that Bloop incurs a negligible training cost compared to the standard regularized training, as it only requires two additional dot products in the parameter space.

The code for the Bloop method is available at \url{https://github.com/apple/ml-bloop}.

\subsection{Baselines and evaluation}
We compare Bloop (\cref{alg:bloop}) to other popular gradient surgery methods that follow a similar design.
We focus on the stochastic setup where we only have access to the gradients over a mini-batch of samples at each iteration.

\textbf{Mixed.} This method minimizes the regularized objective $\lmain+\lambda \laux$ with the direction $d  = g^{\mathrm{batch}}_\main + \lambda g^{\mathrm{batch}}_\aux$.

\textbf{Dynamic Barrier (DB).} 
The original formulation of the DB method requires both an estimate of a lower bound on $\lmain$, as well as an estimate of $\lmain(\theta)$, which are cumbersome to estimate in deep learning setups.
We therefore forgo this part of the algorithm and instead incorporate the scaling factor $\lambda$ to control the trade-off. We also replace the gradients in the original method by stochastic gradients. This results in the update direction $d = \mu  g^{\mathrm{batch}}_\main + \lambda g^{\mathrm{batch}}_\aux$ where
$\mu = \max\left(1 - \lambda\frac{\langle g^{\mathrm{batch}}_\main,g^{\mathrm{batch}}_\aux \rangle }{\|g^{\mathrm{batch}}_\main\|^2}, 0\right).$

\textbf{PCGrad.} Being motivated from a multi-task perspective, the original formulation of PCGrad does not use the scaling factor $\lambda$.
By incorporating this factor, the update direction becomes $d = g^{\mathrm{batch}}_\main + \lambda g^{\mathrm{batch}}_\aux$ if $\langle g^{\mathrm{batch}}_\main, g^{\mathrm{batch}}_\aux\rangle >0$, and $d = \pi(g^{\mathrm{batch}}_\main, g^{\mathrm{batch}}_\aux) + \lambda \pi(g^{\mathrm{batch}}_\aux, g^{\mathrm{batch}}_\main)$ otherwise.

\textbf{Evaluation of the algorithms.}
To provide a comprehensive insight into how the algorithm design affects the training dynamics, we report the metrics on both the training and the test sets. Moreover, we trace the evolution of these metrics along training.

\textbf{Pareto fronts.}
All algorithms that we consider here have thus a parameter $\lambda$ that trades-off between the train and the auxiliary losses. 
After a fixed number of iterations, the algorithm $\texttt{algo}$ finds a final parameter $\theta^{\texttt{algo}}(\lambda)$ that explicitly depends on $\lambda$.
Generally, $\lmain(\theta^{\texttt{algo}}(\lambda))$ is a decreasing function of $\lambda$ while  $\laux(\theta^{\texttt{algo}}(\lambda))$ is increasing with $\lambda$.
We can then vary $\lambda$ to get the set of pairs $\mathcal{P}(\texttt{algo})=\{(\lmain(\theta^{\texttt{algo}}(\lambda)), \laux((\theta^{\texttt{algo}}(\lambda)))|\enspace \lambda \geq 0\}$, called the Pareto front of $\texttt{algo}$.

\subsection{Imposing an explicit bias during training}

To begin with, we first investigate the situation where the auxiliary objective is used to enforce a certain desirable property (bias) on the neural network.

\textbf{Training smooth neural networks.}\enspace
Following our discussion in \cref{sec:intro}, we explore the potential of Bloop in training smooth neural networks.
For this, we use the MNIST dataset~\cite{lecun2010mnist} and an MLP of two hidden layers.
With this minimal architecture, a simple induction argument shows that the Lipschitz constant of the network is upper-bounded by $\prod_{l=1}^L \|W_l\|_2$, where $W_l$ is the weight matrix of the $l-$th linear layer, $\|\cdot\|_2$ is the spectral norm, and $L=3$ is the number of layers. We thus define the auxiliary loss as $\laux=\log(\prod_{l=1}^L \|W_l\|_2)$. The use of logarithm here makes training easier. 
On the other hand, we use the standard cross-entropy loss as the main loss.



\textbf{Training networks with small weights.}\enspace For this experiment, we train a ResNet50 using standard cross-entropy loss on Imagenet, and try to simultaneously achieve a low $\ell_2$ norm of the parameters of the network.
The auxiliary loss is therefore $\laux(\theta) =\frac12\|\theta\|^2$.
In that case, the mixed method is similar to training with a weight decay $\lambda$.

\textbf{Results.}
The results are reported in \cref{fig:exp-bias}.
We see Bloop induces training trajectories that are fundamentally different from all other methods, and leads to better Pareto fronts when trading off the main and the auxiliary training losses.
In both experiments, we observe that Bloop leads to a significantly better Pareto front when looking at the training loss (\autoref{fig:lipschitz}, left and \autoref{fig:imagenet_l2}, left). Whether this translates or not to a better Pareto front in terms of test loss is problem dependent: in \autoref{fig:lipschitz}, right, the Pareto front of Bloop is only slightly better than that of the other methods, while in \autoref{fig:imagenet_l2}, right, it is significantly better.

\begin{figure}[t]
    \centering
\includegraphics[width=.48\columnwidth]{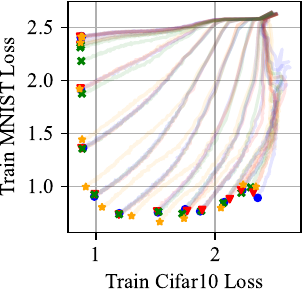}
\hfill
\includegraphics[width=.48\columnwidth]{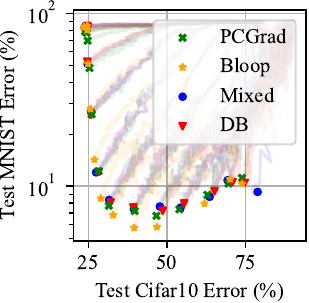}
    \caption{Trade-off between the performances in the Cifar10Mnist multi-task learning problem. Bloop gives a better Pareto front.}
\label{fig:cifar_mnist}
\vspace{-0.5cm}
\end{figure}

\begin{figure*}[t]
    \centering
    \begin{subfigure}[t]{0.49\textwidth}
\includegraphics[width=.495\columnwidth]{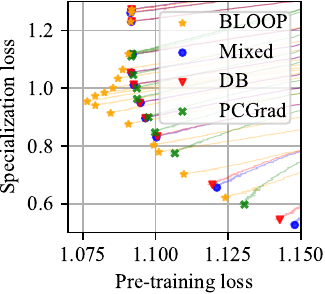}
\hfill
\includegraphics[width=.48\columnwidth]{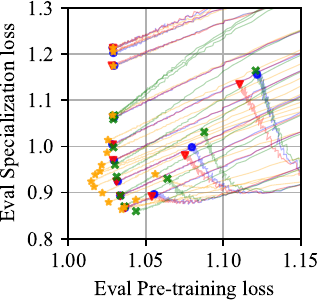}
    \caption{Results on the language modeling task.
    The main, pre-training loss is the next-token-prediction loss over the large c4 dataset, while the auxiliary, specialization loss is the next-token-prediction loss over the small RCV-1 dataset.}
\label{fig:nlp_pretrain}
    \end{subfigure}
    \hfill
    \begin{subfigure}[t]{0.49\textwidth}
    \includegraphics[width=.48\columnwidth]{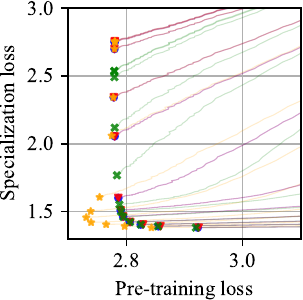}
    \hfill
\includegraphics[width=.5\columnwidth]{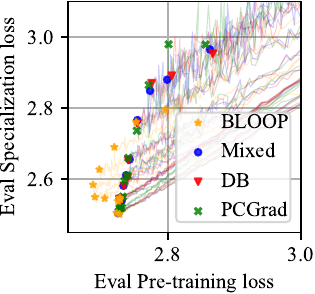}
    \caption{Results on the translation task.
    The main pre-training loss is the translation loss over the large paracrawl dataset, while the auxiliary specialization loss is the translation loss over the small WMT dataset.}
\label{fig:translation}
    \end{subfigure}
    \caption{Trade-offs between the main and the auxiliary objectives in problems in natural language processing experiments with transformer models, where the main loss is the loss over a large dataset and the auxiliary loss is a loss over a small dataset that can be overfitted easily. We observe that Bloop gets a significantly better Pareto front than all other methods, which perform similarly to the mixed method.
    Bloop gains in terms of optimization on the training losses transfer to the evaluation losses.
    }
    \label{fig:exp-joint}
\end{figure*}
\subsection{Multi-task learning}
As discussed in \cref{sec:intro}, multi-task learning represents another typical scenario in which such auxiliary objectives emerge.
Following \citet{hotegni2023multi}, we construct a Cifar10Mnist dataset by overlapping digits from MNIST on images from CIFAR-10~\cite{krizhevsky2009learning} --- see \cref{fig:cifar10mnist} in \cref{apx:exp} for an illustration.
The main and the auxiliary tasks correpond respectively to identifying the label for the background CIFAR-10 image and for the MNIST digit. There is a natural hierachy between the two tasks here because identifying the CIFAR-10 label is more difficult than identifying the MNIST one.
For this dataset, we train a ResNet18 with two classification heads to minimize the two cross-entropy losses. In this experiment, we found that taking $\rho=0.001$ for Bloop gave better results.

\textbf{Results.}\enspace
As shown in \cref{fig:cifar_mnist}, the trajectories of Bloop are again much more different than those of the other methods, which share quite similar behaviors.
Moreover, Bloop gets a slightly improved Pareto front over those methods.

\subsection{Joint training on two datasets}

\begin{figure*}[t]
    \centering
\includegraphics[width=.98\textwidth]{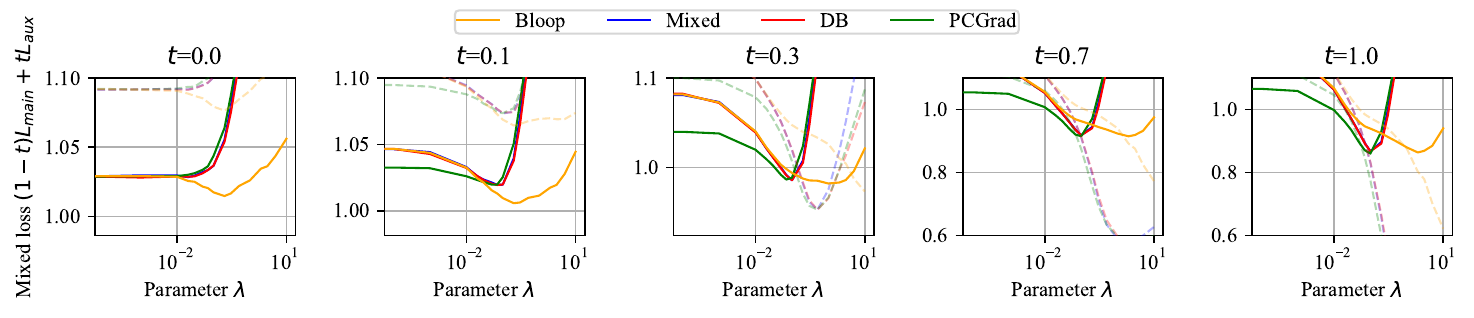}
    \caption{A different look at the results in \autoref{fig:nlp_pretrain}.
    We display the value of the final mixed loss $(1-t)L_\main + t L_\aux$ for the different values of $\lambda$ in the algorithms we used. Bold lines correspond to evaluation loss, while dotted lines correspond to train loss.
    We see that Bloop allows to get to a lower mixed loss when $t$ is small. This is a striking phenomenon, since the mixed method directly minimizes that loss.}
\label{fig:1d_plot}
\end{figure*}

With the advent of large foundation models, it becomes increasingly common to train a model on multiple data sources \cite{gunasekar2023textbooks,sun2023generative,xu2023demystifying,oquab2024dinov}.
Yet, these datasets could have intrinsically different characteristics, and it may be natural to prioritize one over another, for instance when one dataset has far more samples than another.
We explore the benefit of Bloop in such multi-dataset setting.
Our experimental setup is similar to that of \citet{grangier2023adaptive}.

\textbf{Transformer pre-training.}\enspace
We consider the problem of performing next-token-prediction with a decoder-only transformer on text data.
The network is a transformer with 12 decoder layers, 8 attention heads, a residual
dimension of 256, and a feed-forward latent dimension of 1024.
The main loss corresponds to the prediction loss over a large pre-training dataset, while the auxiliary loss corresponds to that on a smaller but higher-quality dataset. 
Due to the lack of data, training only on the small high-quality dataset leads to severe overfitting and poor performance; hence, we resort to training on both datasets, using the proposed baselines or Bloop.
For the training set, we use $30$M examples from the c4 dataset~\citep{raffel2020exploring}, while the auxiliary loss corresponds to $20$K examples from the RCV-1 dataset~\citep{lewis2004rcv1}.

\textbf{Translation.}\enspace
In this experiment, we train a network to translate English into German. 
The network is a transformer with 6 encoder layers and 6 decoder layers, 16 attention heads, a residual
dimension of 1,024, and a feed-forward latent dimension of 4,096.
Like in the pre-training experiment, we have a large generic dataset, the Paracrawl dataset~\citep{banon2020paracrawl}, with 36m sentence pairs, which defines the main loss. 
The auxiliary loss is the loss over a smaller but higher quality dataset, the 2009-2019 WMT dataset, yielding 10k sentence pairs~\citep{farhad2021findings}.
We use the 2020 WMT dataset (2k pairs) as an evaluation set.

\textbf{Results.}\enspace
\autoref{fig:exp-joint} displays the results. 
We observe siginificantly improved results for Bloop, which has once again a better Pareto front, and achieves smaller pre-training loss. 
These gains are kept when looking at the evaluation losses.
\autoref{fig:1d_plot} gives a different perspective on those results.

\begin{figure}[t]
    \centering
\includegraphics[width=.49\columnwidth]{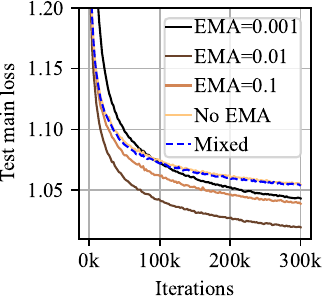}
\includegraphics[width=.48\columnwidth]{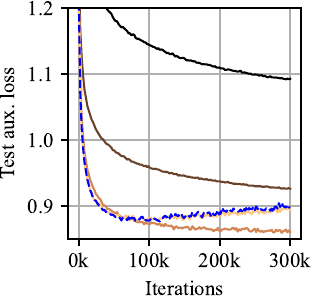}
    \caption{Effect of the EMA parameter $\rho$ on Bloop's performance. 
    We use the same next-token prediction losses as in \autoref{fig:nlp_pretrain}, and display the training curves for a fixed $\lambda=0.2$.}
\label{fig:ema_effect}
\vspace{-0.5cm}
\end{figure}

\subsection{Role of the EMA}
\label{subsec:expe_ema}
We investigate the importance of the EMA parameter $\rho$ in Bloop.
As already seen in \autoref{sec:theory}, it is critical from a theoretical point-of-view for the algorithm's convergence. 
We further illustrate this via the transformer pre-training experiment with a fixed $\lambda=0.2$.

\autoref{fig:ema_effect} displays the results. 
We see that when the EMA is too small ($\rho=0.001$), the value of $g_\main^{\mathrm{EMA}}$ is outdated compared to the current value of the gradient $g_\main$, and therefore, the performance on both the main and auxiliary losses is bad.
On the contrary, taking a too-large EMA ($\rho=0.9$) means that $g_\main^{\mathrm{EMA}}$ has a high variance, and we recover a trajectory extremely similar to that of the mixed method.
Choices between these two extremes ($\rho=0.01$, or $\rho=0.1$) lead to a tradeoff between main and auxiliary loss. 


\section*{Discussion}
A striking phenomenon that we observe in all our experiments is that PCGrad and DB work very similarly to the mixed method. 
We posit that this observation is due to the high gradient variance coming from the main loss, which is also what our theory predicts.
Adding an EMA to reduce this variance leads to the Bloop method, which here has a different behavior to the other methods, often leading to improved Pareto fronts.

In the Appendix~\ref{app:sec:extra}, we describe an experiment where Bloop does not work better than the other methods. We attempted to train a ResNet to have a good performance on Imagenet and Cifar10, with a shared trunk and two classification heads. 
We found that all methods performed equally well; in that case, Bloop leads to the same Pareto front as the other method. 
Yet, once again, PCGrad and DB have the same practical performance as the mixed method.

Overall, adding an EMA to reduce variance in the projection direction is a simple idea that can have a big impact on gradient surgery methods.
\section*{Acknowledgements}
The authors thank Alaa El Nouby, David Grangier, Miguel Sarabia del Castillo, Arno Blaas, Jason Ramapuram, Dan Busbridge, Adam Golinski, Luca Zappella and Federico Danielli for fruitful discussions.
The authors are indebted to David Grangier and Awni Hannun for their help with the codebase.
\section*{Impact Statement}
This paper presents work whose goal is to advance the field of Machine Learning. There are many potential societal consequences of our work, none which we feel must be specifically highlighted here.
\bibliography{references}

\begin{thebibliography}{35}
\providecommand{\natexlab}[1]{#1}
\providecommand{\url}[1]{\texttt{#1}}
\expandafter\ifx\csname urlstyle\endcsname\relax
  \providecommand{\doi}[1]{doi: #1}\else
  \providecommand{\doi}{doi: \begingroup \urlstyle{rm}\Url}\fi

\bibitem[Ba{\~n}{\'o}n et~al.(2020)Ba{\~n}{\'o}n, Chen, Haddow, Heafield,
  Hoang, Espl{\`a}-Gomis, Forcada, Kamran, Kirefu, Koehn,
  et~al.]{banon2020paracrawl}
Ba{\~n}{\'o}n, M., Chen, P., Haddow, B., Heafield, K., Hoang, H.,
  Espl{\`a}-Gomis, M., Forcada, M., Kamran, A., Kirefu, F., Koehn, P., et~al.
\newblock Paracrawl: Web-scale acquisition of parallel corpora.
\newblock Association for Computational Linguistics (ACL), 2020.

\bibitem[Cao et~al.(2023)Cao, Jiang, Abolfazli, Hamedani, and
  Mokhtari]{cao2023projection}
Cao, J., Jiang, R., Abolfazli, N., Hamedani, E.~Y., and Mokhtari, A.
\newblock Projection-free methods for stochastic simple bilevel optimization
  with convex lower-level problem.
\newblock \emph{arXiv preprint arXiv:2308.07536}, 2023.

\bibitem[Caruana(1997)]{caruana1997multitask}
Caruana, R.
\newblock Multitask learning.
\newblock \emph{Machine learning}, 28:\penalty0 41--75, 1997.

\bibitem[Chaudhry et~al.(2018)Chaudhry, Ranzato, Rohrbach, and
  Elhoseiny]{chaudhry2018efficient}
Chaudhry, A., Ranzato, M., Rohrbach, M., and Elhoseiny, M.
\newblock Efficient lifelong learning with a-gem.
\newblock In \emph{International Conference on Learning Representations}, 2018.

\bibitem[Cisse et~al.(2017)Cisse, Bojanowski, Grave, Dauphin, and
  Usunier]{cisse2017parseval}
Cisse, M., Bojanowski, P., Grave, E., Dauphin, Y., and Usunier, N.
\newblock Parseval networks: Improving robustness to adversarial examples.
\newblock In \emph{International conference on machine learning}, pp.\
  854--863. PMLR, 2017.

\bibitem[Cooper(2018)]{cooper2018loss}
Cooper, Y.
\newblock The loss landscape of overparameterized neural networks.
\newblock \emph{arXiv preprint arXiv:1804.10200}, 2018.

\bibitem[DeepMind et~al.(2020)DeepMind, Babuschkin, Baumli, Bell, Bhupatiraju,
  Bruce, Buchlovsky, Budden, Cai, Clark, Danihelka, Dedieu, Fantacci, Godwin,
  Jones, Hemsley, Hennigan, Hessel, Hou, Kapturowski, Keck, Kemaev, King,
  Kunesch, Martens, Merzic, Mikulik, Norman, Papamakarios, Quan, Ring, Ruiz,
  Sanchez, Sartran, Schneider, Sezener, Spencer, Srinivasan, Stanojevi\'{c},
  Stokowiec, Wang, Zhou, and Viola]{deepmind2020jax}
DeepMind, Babuschkin, I., Baumli, K., Bell, A., Bhupatiraju, S., Bruce, J.,
  Buchlovsky, P., Budden, D., Cai, T., Clark, A., Danihelka, I., Dedieu, A.,
  Fantacci, C., Godwin, J., Jones, C., Hemsley, R., Hennigan, T., Hessel, M.,
  Hou, S., Kapturowski, S., Keck, T., Kemaev, I., King, M., Kunesch, M.,
  Martens, L., Merzic, H., Mikulik, V., Norman, T., Papamakarios, G., Quan, J.,
  Ring, R., Ruiz, F., Sanchez, A., Sartran, L., Schneider, R., Sezener, E.,
  Spencer, S., Srinivasan, S., Stanojevi\'{c}, M., Stokowiec, W., Wang, L.,
  Zhou, G., and Viola, F.
\newblock The {D}eep{M}ind {JAX} {E}cosystem, 2020.
\newblock URL \url{http://github.com/google-deepmind}.

\bibitem[Dempe et~al.(2010)Dempe, Dinh, and Dutta]{dempe2010optimality}
Dempe, S., Dinh, N., and Dutta, J.
\newblock Optimality conditions for a simple convex bilevel programming
  problem.
\newblock \emph{Variational Analysis and Generalized Differentiation in
  Optimization and Control: In Honor of Boris S. Mordukhovich}, pp.\  149--161,
  2010.

\bibitem[Dery et~al.(2021)Dery, Dauphin, and Grangier]{dery2021auxiliary}
Dery, L.~M., Dauphin, Y., and Grangier, D.
\newblock Auxiliary task update decomposition: The good, the bad and the
  neutral.
\newblock \emph{arXiv preprint arXiv:2108.11346}, 2021.

\bibitem[Du et~al.(2018)Du, Czarnecki, Jayakumar, Farajtabar, Pascanu, and
  Lakshminarayanan]{du2018adapting}
Du, Y., Czarnecki, W.~M., Jayakumar, S.~M., Farajtabar, M., Pascanu, R., and
  Lakshminarayanan, B.
\newblock Adapting auxiliary losses using gradient similarity.
\newblock \emph{arXiv preprint arXiv:1812.02224}, 2018.

\bibitem[Farhad et~al.(2021)Farhad, Arkady, Magdalena, Ond{\v{r}}ej, Rajen,
  Vishrav, Costa-jussa, Cristina, Angela, Christian,
  et~al.]{farhad2021findings}
Farhad, A., Arkady, A., Magdalena, B., Ond{\v{r}}ej, B., Rajen, C., Vishrav,
  C., Costa-jussa, M.~R., Cristina, E.-B., Angela, F., Christian, F., et~al.
\newblock Findings of the 2021 conference on machine translation (wmt21).
\newblock In \emph{Proceedings of the Sixth Conference on Machine Translation},
  pp.\  1--88. Association for Computational Linguistics, 2021.

\bibitem[Gong \& Liu(2021)Gong and Liu]{gong2021bi}
Gong, C. and Liu, X.
\newblock Bi-objective trade-off with dynamic barrier gradient descent.
\newblock \emph{NeurIPS 2021}, 2021.

\bibitem[Grangier et~al.(2023)Grangier, Ablin, and
  Hannun]{grangier2023adaptive}
Grangier, D., Ablin, P., and Hannun, A.
\newblock Adaptive training distributions with scalable online bilevel
  optimization.
\newblock \emph{arXiv preprint arXiv:2311.11973}, 2023.

\bibitem[Gunasekar et~al.(2023)Gunasekar, Zhang, Aneja, Mendes, Del~Giorno,
  Gopi, Javaheripi, Kauffmann, de~Rosa, Saarikivi,
  et~al.]{gunasekar2023textbooks}
Gunasekar, S., Zhang, Y., Aneja, J., Mendes, C. C.~T., Del~Giorno, A., Gopi,
  S., Javaheripi, M., Kauffmann, P., de~Rosa, G., Saarikivi, O., et~al.
\newblock Textbooks are all you need.
\newblock \emph{arXiv preprint arXiv:2306.11644}, 2023.

\bibitem[He et~al.(2022)He, Feng, Cheng, Ji, Guo, and
  Caverlee]{he2022metabalance}
He, Y., Feng, X., Cheng, C., Ji, G., Guo, Y., and Caverlee, J.
\newblock Metabalance: improving multi-task recommendations via adapting
  gradient magnitudes of auxiliary tasks.
\newblock In \emph{Proceedings of the ACM Web Conference 2022}, pp.\
  2205--2215, 2022.

\bibitem[Heek et~al.()Heek, Levskaya, Oliver, Ritter, Rondepierre, Steiner, and
  van Zee]{heek1flax}
Heek, J., Levskaya, A., Oliver, A., Ritter, M., Rondepierre, B., Steiner, A.,
  and van Zee, M.
\newblock Flax: A neural network library and ecosystem for jax, 2020.
\newblock \emph{URL http://github. com/google/flax}, 1.

\bibitem[Hotegni et~al.(2023)Hotegni, Berkemeier, and Peitz]{hotegni2023multi}
Hotegni, S.~S., Berkemeier, M., and Peitz, S.
\newblock Multi-objective optimization for sparse deep multi-task learning.
\newblock \emph{arXiv preprint arXiv:2308.12243}, 2023.

\bibitem[Karimi et~al.(2016)Karimi, Nutini, and Schmidt]{karimi2016linear}
Karimi, H., Nutini, J., and Schmidt, M.
\newblock Linear convergence of gradient and proximal-gradient methods under
  the polyak-{\l}ojasiewicz condition.
\newblock In \emph{Machine Learning and Knowledge Discovery in Databases:
  European Conference, ECML PKDD 2016, Riva del Garda, Italy, September 19-23,
  2016, Proceedings, Part I 16}, pp.\  795--811. Springer, 2016.

\bibitem[Kingma \& Ba(2014)Kingma and Ba]{kingma2014adam}
Kingma, D.~P. and Ba, J.
\newblock Adam: A method for stochastic optimization.
\newblock \emph{arXiv preprint arXiv:1412.6980}, 2014.

\bibitem[Krizhevsky et~al.(2009)Krizhevsky, Hinton,
  et~al.]{krizhevsky2009learning}
Krizhevsky, A., Hinton, G., et~al.
\newblock Learning multiple layers of features from tiny images.
\newblock 2009.

\bibitem[LeCun et~al.(2010)LeCun, Cortes, and Burges]{lecun2010mnist}
LeCun, Y., Cortes, C., and Burges, C.
\newblock Mnist handwritten digit database.
\newblock \emph{ATT Labs [Online]. Available:
  http://yann.lecun.com/exdb/mnist}, 2, 2010.

\bibitem[Lewis et~al.(2004)Lewis, Yang, Russell-Rose, and Li]{lewis2004rcv1}
Lewis, D.~D., Yang, Y., Russell-Rose, T., and Li, F.
\newblock Rcv1: A new benchmark collection for text categorization research.
\newblock \emph{Journal of machine learning research}, 5\penalty0
  (Apr):\penalty0 361--397, 2004.

\bibitem[Li et~al.(2018)Li, Xu, Taylor, Studer, and
  Goldstein]{li2018visualizing}
Li, H., Xu, Z., Taylor, G., Studer, C., and Goldstein, T.
\newblock Visualizing the loss landscape of neural nets.
\newblock \emph{Advances in neural information processing systems}, 31, 2018.

\bibitem[Liu et~al.(2022)Liu, Zhu, and Belkin]{liu2022loss}
Liu, C., Zhu, L., and Belkin, M.
\newblock Loss landscapes and optimization in over-parameterized non-linear
  systems and neural networks.
\newblock \emph{Applied and Computational Harmonic Analysis}, 59:\penalty0
  85--116, 2022.

\bibitem[Luo \& Tseng(1993)Luo and Tseng]{luo1993error}
Luo, Z.-Q. and Tseng, P.
\newblock Error bounds and convergence analysis of feasible descent methods: a
  general approach.
\newblock \emph{Annals of Operations Research}, 46\penalty0 (1):\penalty0
  157--178, 1993.

\bibitem[Oquab et~al.(2024)Oquab, Darcet, Moutakanni, Vo, Szafraniec, Khalidov,
  Fernandez, HAZIZA, Massa, El-Nouby, Assran, Ballas, Galuba, Howes, Huang, Li,
  Misra, Rabbat, Sharma, Synnaeve, Xu, Jegou, Mairal, Labatut, Joulin, and
  Bojanowski]{oquab2024dinov}
Oquab, M., Darcet, T., Moutakanni, T., Vo, H.~V., Szafraniec, M., Khalidov, V.,
  Fernandez, P., HAZIZA, D., Massa, F., El-Nouby, A., Assran, M., Ballas, N.,
  Galuba, W., Howes, R., Huang, P.-Y., Li, S.-W., Misra, I., Rabbat, M.,
  Sharma, V., Synnaeve, G., Xu, H., Jegou, H., Mairal, J., Labatut, P., Joulin,
  A., and Bojanowski, P.
\newblock {DINO}v2: Learning robust visual features without supervision.
\newblock \emph{Transactions on Machine Learning Research}, 2024.
\newblock ISSN 2835-8856.
\newblock URL \url{https://openreview.net/forum?id=a68SUt6zFt}.

\bibitem[Raffel et~al.(2020)Raffel, Shazeer, Roberts, Lee, Narang, Matena,
  Zhou, Li, and Liu]{raffel2020exploring}
Raffel, C., Shazeer, N., Roberts, A., Lee, K., Narang, S., Matena, M., Zhou,
  Y., Li, W., and Liu, P.~J.
\newblock Exploring the limits of transfer learning with a unified text-to-text
  transformer.
\newblock \emph{The Journal of Machine Learning Research}, 21\penalty0
  (1):\penalty0 5485--5551, 2020.

\bibitem[Sabach \& Shtern(2017)Sabach and Shtern]{sabach2017first}
Sabach, S. and Shtern, S.
\newblock A first order method for solving convex bilevel optimization
  problems.
\newblock \emph{SIAM Journal on Optimization}, 27\penalty0 (2):\penalty0
  640--660, 2017.

\bibitem[Sener \& Koltun(2018)Sener and Koltun]{sener2018multi}
Sener, O. and Koltun, V.
\newblock Multi-task learning as multi-objective optimization.
\newblock \emph{Advances in neural information processing systems}, 31, 2018.

\bibitem[Sun et~al.(2023)Sun, Cui, Zhang, Zhang, Yu, Luo, Wang, Rao, Liu,
  Huang, et~al.]{sun2023generative}
Sun, Q., Cui, Y., Zhang, X., Zhang, F., Yu, Q., Luo, Z., Wang, Y., Rao, Y.,
  Liu, J., Huang, T., et~al.
\newblock Generative multimodal models are in-context learners.
\newblock \emph{arXiv preprint arXiv:2312.13286}, 2023.

\bibitem[Terj{\'e}k(2019)]{terjek2019adversarial}
Terj{\'e}k, D.
\newblock Adversarial lipschitz regularization.
\newblock \emph{arXiv preprint arXiv:1907.05681}, 2019.

\bibitem[Tsuzuku et~al.(2018)Tsuzuku, Sato, and Sugiyama]{tsuzuku2018lipschitz}
Tsuzuku, Y., Sato, I., and Sugiyama, M.
\newblock Lipschitz-margin training: Scalable certification of perturbation
  invariance for deep neural networks.
\newblock \emph{Advances in neural information processing systems}, 31, 2018.

\bibitem[Wang \& Tsvetkov(2021)Wang and Tsvetkov]{wang2021gradient}
Wang, Z. and Tsvetkov, Y.
\newblock Gradient vaccine: Investigating and improving multi-task optimization
  in massively multilingual models.
\newblock In \emph{Proceedings of the International Conference on Learning
  Representations (ICLR)}, 2021.

\bibitem[Xu et~al.(2023)Xu, Xie, Tan, Huang, Howes, Sharma, Li, Ghosh,
  Zettlemoyer, and Feichtenhofer]{xu2023demystifying}
Xu, H., Xie, S., Tan, X.~E., Huang, P.-Y., Howes, R., Sharma, V., Li, S.-W.,
  Ghosh, G., Zettlemoyer, L., and Feichtenhofer, C.
\newblock Demystifying clip data.
\newblock \emph{arXiv preprint arXiv:2309.16671}, 2023.

\bibitem[Yu et~al.(2020)Yu, Kumar, Gupta, Levine, Hausman, and
  Finn]{yu2020gradient}
Yu, T., Kumar, S., Gupta, A., Levine, S., Hausman, K., and Finn, C.
\newblock Gradient surgery for multi-task learning.
\newblock \emph{Advances in Neural Information Processing Systems},
  33:\penalty0 5824--5836, 2020.

\end{thebibliography}
\bibliographystyle{icml2024}

\newpage
\appendix
\onecolumn
\noindent\rule{\textwidth}{1pt}
\begin{center}
\vspace{7pt}
{\Large \fontseries{bx}\selectfont Appendix}
\end{center}
\noindent\rule{\textwidth}{1pt}

\section{Convergence analysis}
\label{apx:proofs}
In this appendix we provide proofs for the theoretical results of \cref{sec:theory}.

\subsection{Proof of \autoref{prop:stat}}

In the following, we will prove the two implications in the proposition separately.

\paragraph{Small Bloop Vector $\rightarrow$ Near-Stationary Point.}
By orthogonality, we have  $\|d\|^2 = \|g_\main\|^2 + \lambda^2 \|\pi(g_\aux; g_\main)\|^2$.
This implies immediately $\|g_\main\|\leq \varepsilon$ and $\|\pi(g_\aux; g_\main)\|\leq \varepsilon\lambda^{-1}$ provided that $\|d\|\leq\varepsilon$.

Let us next consider the case where \cref{assum:local_error_bound} holds and that the Hessian of $\lmain$ is M-Lipschitz continuous.
With the local error bound, i.e.,  \cref{assum:local_error_bound}, we know there exists $\theta^*$ such that $\nabla \lmain(\theta^*) = 0$ and $\|\theta - \theta^*\|\leq c \|g_\main\|$.
Performing a Taylor expansion with Lagrange form of the remainder of order 2, we obtain
\begin{equation}
\begin{aligned}[b]
\label{eq:taylor-order-2}
\nabla L_{\text{main}}(\theta) 
& =  \nabla \lmain(\theta^{\ast}) +  \nabla^{2} L_{\text{main}}(\theta^{\ast}) (\theta-\theta^{\ast}) + \frac{1}{2}\nabla^{3} L_{\text{main}}(\theta^{\prime})[\theta-\theta^{\ast}, \theta-\theta^{\ast}]\\
& =\nabla^{2} L_{\text{main}}(\theta^{\ast}) (\theta-\theta^{\ast}) + \frac{1}{2}\nabla^{3} L_{\text{main}}(\theta^{\prime})[\theta-\theta^{\ast}, \theta-\theta^{\ast}],
\end{aligned}
\end{equation}
for some $\theta^{\prime}$ that lies on the line that connects $\theta$ and $\theta^{\ast}$.
Using the M-Lipschitzness of $\nabla^2\lmain$, the norm of 
$r = \nabla \lmain(\theta) - \nabla^2\lmain(\theta)(\theta-\theta^*)$ can then be bounded by
$$
\|r\| = \frac{1}{2}\nabla^{3} L_{\text{main}}(\theta^{\prime})[\theta-\theta^{\ast}, \theta-\theta^{\ast}] \leq \frac{M}{2}\|\theta - \theta^*\|^2\leq \frac{Mc^2}{2}\|g_\main\|^2\enspace.
$$
We now claim that the desired inequality holds true with
\[
v =\frac{\langle g_\aux, g_\main\rangle}{\|g_\main\|^2}(\theta - \theta^*).
\]
For this, we decompose
$$
\frac{\langle g_\aux, g_\main\rangle}{\|g_\main\|^2}g_\main = \nabla^2\lmain(\theta)v + \frac{\langle g_\aux, g_\main\rangle}{\|g_\main\|^2}r
$$
Subsequently,
\begin{align*}
    \|g_\aux - \nabla^2L_\main(\theta)v\| &\leq 
    \left\|g_\aux - \frac{\langle g_\aux, g_\main\rangle}{\|g_\main\|^2}g_\main\right\| + \left\|\frac{\langle g_\aux, g_\main\rangle}{\|g_\main\|^2}r\right\|\\
    &\leq \|\pi(g_\aux; g_\main)\| + \frac{Mc^2\|g_\aux\|\|g_\main\|}{2}\\
    &\leq \left(\lambda^{-1} + \frac{Mc^2}{2} \|g_\aux\|\right)\varepsilon.
\end{align*}

\paragraph{Near-Stationary Point $\rightarrow$ Small Bloop Vector.}

Reciprocally, by plugging $\theta=\theta^* + \varepsilon v$ into \eqref{eq:taylor-order-2}, we get
\begin{equation*}
g_\main = \nabla \lmain(\theta) 
= \varepsilon \nabla^{2} L_{\text{main}}(\theta^{\ast}) v + \frac{\varepsilon^2}{2}\nabla^{3} L_{\text{main}}(\theta^{\prime})[v, v]
= \varepsilon \nabla^{2} L_{\text{main}}(\theta^{\ast}) v
+ o(\varepsilon).
\end{equation*}
Moreover, by continuity of $\nabla \laux$ and the optimality condition $\nabla \laux(\theta^{\ast})=\nabla^2\lmain(\theta)v$, we have $ g_\aux = \nabla^2\lmain(\theta^{\ast})v + o(1)$ when $\varepsilon$ goes to $0$.


From here, we will prove $\lim_{\varepsilon\rightarrow0}\pi(g_\aux;g_\main)=0$ by distinguishing between two cases:


\textbf{Case 1: $L_{\text{aux}}(\theta^{\ast})=\mathbf{0}$.} \quad
In other words, $\nabla^2 L_{\text{main}}(\theta^{\ast})v=\mathbf{0}$. Thus $g_{\text{aux}}=o(1)$. $\pi(g_{\text{aux}};g_{\text{main}})$ being a projection of $g_{\text{aux}}$, we have $\|\pi(g_{\text{aux}};g_{\text{main}})\|\le\|g_{\text{aux}}\|$. This show $\pi(g_{\text{aux}};g_{\text{main}})=o(1)$.


\textbf{Case 2: $\nabla L_{\text{aux}}(\theta^{\ast})\neq\mathbf{0}$.}\quad
This indicates $\nabla^2 L_{\text{main}}(\theta^{*})v\neq\mathbf{0}$.
We use the formula
\begin{equation}
\label{eq:bloop-proj-proof}
\pi(g_{\text{aux}}; g_{\text{main}}) = g_{\text{aux}} - \frac{\langle g_{\text{aux}}, g_{\text{main}} \rangle}{\|g_{\text{main}}\|^2}g_{\text{main}}. 
\end{equation}
With $\langle g_{\text{aux}}, g_{\text{main}} \rangle = \varepsilon \|\nabla^2 L_{\text{main}}(\theta^{\ast})v\|^2 + o(\varepsilon)$ and $\|g_{\text{main}}\|^2=\varepsilon^2 \|\nabla^2 L_{\text{main}}(\theta^{\ast})v\|^2 + o(\varepsilon^2)$, we have
$$\frac{\varepsilon\langle g_{\text{aux}}, g_{\text{main}} \rangle}{\|g_{\text{main}}\|^2} = \frac{\|\nabla^2 L_{\text{main}}(\theta^{\ast})v\|^2+o(1)}{\|\nabla^2 L_{\text{main}}(\theta^{*})v\|^2+o(1)} = 1+o(1),$$
where the last equality holds since $\|\nabla^2 L_{\text{main}}(\theta^{\ast})v\|^2 \neq 0$.

On the other hand,
$$\frac{g_{\text{main}}}{\varepsilon}=\nabla^2 L_{\text{main}}(\theta^{\ast})v + o(1).$$
We have thus 
$$\frac{\langle g_{\text{aux}}, g_{\text{main}} \rangle}{\|g_{\text{main}}\|^2}g_{\text{main}} = \nabla^2 L_{\text{main}}(\theta^{\ast})v + o(1).$$
Using \eqref{eq:bloop-proj-proof} and $g_\aux = \nabla^2\lmain(\theta^{\ast})v + o(1)$, we deduce $\pi(g_{\text{aux}}; g_{\text{main}})=o(1)$.

\textbf{Conclude.}
In the two cases, we have shown $\lim_{\varepsilon\rightarrow0}\pi(g_{\text{aux}}; g_{\text{main}})=0$. 
Moreover, we also have $\lim_{\varepsilon\rightarrow0}g_\main = 0$.
Adding the two we get exactly $\lim_{\varepsilon\to 0}d(\theta^* + \varepsilon v) = 0$.



\subsection{Proof of \autoref{thm:convergence}}

Here, $\lmain$ and $\laux$ are the empirical risks
$$
\lmain(\theta) =\frac1n \sum_{i=1}^nL_i(\theta)~~\text{ and }~~\laux(\theta) = \frac1m\sum_{j=1}^m L'_j(\theta).
$$
We consider the Bloop method with SGD, which has an EMA $g_{\mathrm{EMA}}^t$ and parameters $\theta^t$ which are updated following
\begin{align*}
    \label{app:eq:update}
    \text{Sample }i, j &\sim \text{ Uniform}\\
    g_{\mathrm{EMA}}^{t+1} &= (1 - \rho)g_{\mathrm{EMA}}^{t} + \rho \nabla L_i(\theta^t)\\
    d^t &=  \nabla L_i(\theta^t) + \lambda \pi( \nabla L'_j(\theta^t);g_{\mathrm{EMA}}^{t})\\
    \theta^{t+1} &= \theta^t - \eta d^t
\end{align*}
Our analysis works by controlling two quantities: the distance from the EMA to the full-batch train gradient
\begin{equation*}
    \label{app:eq:distance_ema}
    \phi_1^t =\mathbb{E}\left[ \|g_{\mathrm{EMA}}^{t+1} - \nabla \lmain(\theta^t)\|^2\right]
\end{equation*}
and the train loss
\begin{equation*}
    \label{app:eq:train_loss}
    \phi_2^t =\mathbb{E}\left[\lmain(\theta^t)\right].
\end{equation*}

\textbf{Control of the EMA.}
For the EMA, we get by expanding
\begin{align*}
    \phi_1^{t+1} &= \mathbb{E}\left[\|g^t_{\mathrm{EMA}} - \rho (g^t_{\mathrm{EMA}} - \nabla L_i(\theta^t)) - \nabla L_{\main}(\theta^{t})\|^2\right]\\
    &=(1-\rho)^2\mathbb{E}\left[\|g^t_{\mathrm{EMA}}  - \nabla L_{\main}(\theta^{t})\|^2\right] + \rho^2\mathbb{E}\left[\|\nabla L_i(\theta^t)-\nabla \lmain(\theta^t)\|^2\right]\\
    &\leq (1 - \rho)\mathbb{E}\left[\|g^t_{\mathrm{EMA}}  - \nabla L_{\main}(\theta^{t})\|^2\right] + \rho^2C^2
\end{align*}

where $C^2$ upper bounds the train gradients variance and where $\rho < 1$.
Let $a= g^t_{\mathrm{EMA}} - \nabla L_{\main}(\theta^{t-1})$ and $b = \nabla L_{\main}(\theta^{t-1}) - \nabla L_{\main}(\theta^{t})$.
Since the inequality $\|a + b\|^2\leq (1 +\delta)\|a\|^2 + (1 +\delta^{-1})\|b\|^2$ holds  true for all $\delta$, we have specifically that
$$
\|g^t_{\mathrm{EMA}}  - \nabla L_{\main}(\theta^{t-1})\|^2 \leq (1 +\delta)\phi_1^t + (1 +\delta^{-1})L^2\eta^2\|d^{t-1}\|^2
$$
for $\delta = \frac\rho 2$. Using $(1 - \rho)(1+\frac\rho2) \leq 1- \frac\rho2$ then gives the descent lemma on the EMA:
\begin{equation*}
    \phi_1^{t+1}\leq \left(1-\frac\rho2\right)\phi_1^t + \rho^2C^2 + \frac{2L^2\eta^2}{\rho}\|d^{t-1}\|^2.
\end{equation*}

Next, we bound crudely $\|d^{t-1}\|\leq D$, and equalize the last two terms, i.e. take $\rho = \left(\frac{2L^2D^2}{C^2}\right)^{\frac13}\eta^{\frac23}$, so that the descent on the EMA becomes 
\begin{equation*}
    \label{app:eq:descent_ema}
    \phi_1^{t+1}\leq \left(1-\frac\rho2\right)\phi_1^t + 2\rho^2C^2
\end{equation*}
which in turn implies that 
$$\phi_1^t\leq 4\rho C^2.$$
\paragraph{Control of the loss.}
The $L$-smoothness of $\lmain$ and the fact that $\mathbb{E}_{i,j}[d^t] = \nabla \lmain(\theta^t) +\lambda \pi(\nabla \laux;  g_{\mathrm{EMA}}^{t})$ gives:
\begin{equation*}
    \phi_2^{t+1}\leq \phi_2^t - \eta \|\nabla \lmain(\theta^t)\|^2 -\eta\lambda \langle \pi(\nabla \laux;  g_{\mathrm{EMA}}^{t}), \nabla \lmain(\theta^t)\rangle +\frac{L\eta^2}2\|d^t\|^2.
\end{equation*}

We omit expectation from the above formula for the ease of presentation, and we will continue doing so for this part of the proof.
The annoying middle term is controlled by 
\begin{align*}
     -\eta\lambda \langle \pi(\nabla \laux;g_{\mathrm{EMA}}^{t}), \nabla \lmain(\theta^t)\rangle &=  -\eta\lambda \langle \pi(\nabla \laux;  g_{\mathrm{EMA}}^{t}), \nabla \lmain(\theta^t) -g_{\mathrm{EMA}}^{t} \rangle \\
     &\leq \eta\lambda B \| \nabla\lmain(\theta^t) -\nabla\lmain(\theta^{t+1})\|
     +
     \eta\lambda B \|\nabla\lmain(\theta^{t+1}) -g_{\mathrm{EMA}}^{t}\| \\
     &\leq 
     \eta^2\lambda LB \| d^t\|
     +
     \eta\lambda B \|\nabla\lmain(\theta^{t+1}) -g_{\mathrm{EMA}}^{t}\|
\end{align*}
where $B$ upper bounds $\|\nabla L_\aux\|$.
The last $\mathbb{E}[\|d^t\|^2]$ is simply bounded by $D^2$.
Hence we get the descent lemma on the train loss:
\begin{equation*}
    \label{app:eq:descent_loss}
    \phi_2^{t+1}\leq \phi_2^t - \eta \|\nabla \lmain(\theta^t)\|^2 +\eta\lambda B\sqrt{\phi_1^t} + \eta^2\left(\frac{LD^2}2 + \lambda LBD\right).
\end{equation*}
Plugging the rate for $\phi_1^t$, we finally get
$$
\phi_2^{t+1}\leq \phi_2^t - \eta \|\nabla \lmain(\theta^t)\|^2 +\eta^{\frac43} C_1 + \eta^2 C_2
$$
for some constants $C_1, C_2 \ge 0$.
In the above we have also used that 
$$\mathbb{E}[\|\nabla\lmain(\theta^{t+1}) -g_{\mathrm{EMA}}^{t}\|]^2 \le \mathbb{E}[\|\nabla\lmain(\theta^{t+1}) -g_{\mathrm{EMA}}^{t}\|^2].$$
Taking 
$\eta \leq \left(\frac{C_1}{C_2}\right)^{\frac32}$
ensures that the last term is smaller than the previous, yielding the simple inequality:
$$
\phi_2^{t+1}\leq \phi_2^t - \eta \|\nabla \lmain(\theta^t)\|^2 +2\eta^{\frac43}C_1.
$$
We now have two kinds of results depending on the context:
\paragraph{Non-convex result.}
Without further assumption, summing the previous inequalities for $t=0\dots T-1$ gives
$$
\frac1T\sum_{t=0}^{T-1}\mathbb{E}[\|\nabla \lmain(\theta^t)\|^2]\leq \frac{\lmain(\theta^0)}{\eta T}+2\eta^{\frac13}C_1.
$$
Hence, taking $\eta \simeq T^{-\frac34}$ gives a $O(T^{-\frac14})$ rate.
\paragraph{PL-result.}
We here assume that $\lmain$ verifies the PL inequality $\frac12\|\nabla \lmain(\theta)\|^2\geq \mu \lmain(\theta)$, where we posit $\min\lmain=0$ without loss of generalitiy.
The descent lemma gives
$$
\phi_2^{t+1}\leq (1 - 2\eta\mu) \phi_2^t+2\eta^{\frac43}C_1.
$$

By unrolling it we obtain
$$
\mathbb{E}[\lmain(\theta^T)]\leq (1 - 2\eta\mu)^T\lmain(\theta^0) + (1 - (1 - 2\eta\mu)^T)\frac{\eta^{\frac13}C_1}{\mu}.
$$
This shows a linear convergence to a radius proportional to $\eta^{\frac13}$.

\section{Experimental Details}
\label{apx:exp}

In this appendix we report the missing details from \cref{sec:expe}.

\textbf{Training smooth networks.}
For this experiment we use an MLP with ReLU activations. The features are of size $728 \rightarrow 256 \rightarrow 128 \rightarrow 10$.
All the methods are trained with Adam optimizer at learning rate of $3\times 10^{-4}$ for 100 epochs and a cosine learning rate schedule. For consistency with the other classification experiments we also include $5$ epochs of warm-up. The batch size is fixed at $256$, and we take a grid of $\lambda$ with $\log_{10}(\lambda) = -4, -3.5, \dots, -0.5, 0$.

\begin{figure}[t]   
    \centering
    \includegraphics[width=0.26\textwidth]{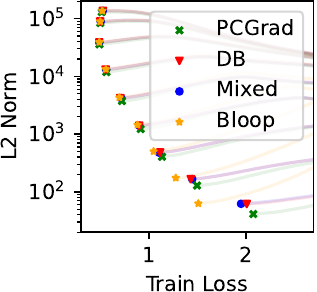}
        \includegraphics[width=0.26\textwidth]{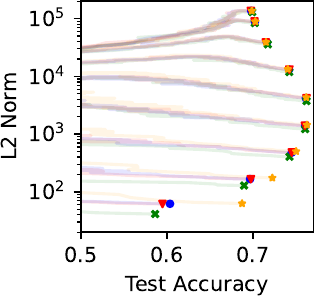}
    \caption{Results of all methods on the imagenet + L2 problem. PCGrad and DB have similar performance to the mixed method.}
    \label{app:fig:imagenet_l2}
\end{figure}

\textbf{Imagenet training with L2 regularization.}
For ImageNet training, we employ SGD with a batch size of $2048$, Nesterov momentum of $0.9$, and a learning rate of $0.8$. This learning rate is derived by scaling the base rate of $0.1$ by a factor of $8$, corresponding to the ratio $2048/256$. Additionally, we apply a cosine learning rate schedule with $5$ warm-up epochs and utilize random cropping and flipping for data augmentation during training. 
The network is trained for $100$ epochs.
This configuration is known to work well for the ResNet50 architecture that we are using here.
The grid of $\lambda$ is  14 uniform values in log scale between $10^{-6}$ and $10^{-2}$, and $0$.
We display results for all methods in \autoref{app:fig:imagenet_l2} with a slightly smaller grid of $\lambda$'s.

\textbf{Multi-task learning with Cifar10Mnist.}
The overall setup for this problem is similar to that for Imagenet training, with the exceptions that we use a smaller architecture---ResNet18 instead of ResNet50, and a smaller batch size---$256$ instead of $2048$.
We also scale down the learning rate to $0.1$ to account for the smaller batch size. The values of the trade-off parameter $\lambda$ goes from $10^{-3}$ to $10^{3}$ and are split equally on log scale.
Unlike Adam, SGD does not adjust the learning rate scale automatically.
This causes unstable training when $\lambda$ is too large. We thus futher scale the learning rate $0.1$ by $1/(1+\lambda)$ for each independent run.

\begin{figure}
    \centering
    \includegraphics[width=0.6\textwidth]{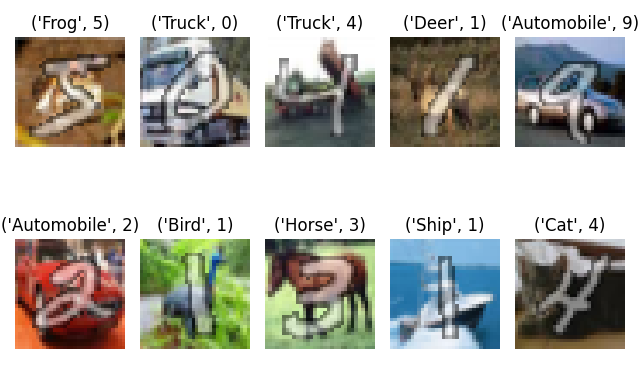}
    \caption{Sample images from the Cifar10Mnist dataset.}
    \label{fig:cifar10mnist}
\end{figure}

\textbf{Next token prediction.} Our model is a byte-level decoder-only transformer. It has 12 layers, 8 attention heads, a residual dimension of 256, and a feed-forward dimension 1024. We use a batch-size of 128 for both datasets, the optimizer is Adam with a learning rate of $0.002$. We train the model for $300K$ iterations.
The grid of $\lambda$ consists of $16$ values evenly spaced in log-space between $10^{-4}$ and $10$, as well as $0$.

\textbf{Translation.} Our model is an encore-decoder transformer. It has 6 encoder and decoder layers, 16 attention heads, a residual dimension of 1024, and a feed-forward dimension 4096. We use a batch-size of 256 for both datasets, the optimizer is Adam with a learning rate of $0.0002$. We train the model for $500K$ iterations. 
Our implementation is derived from the flax example~\citep{heek1flax}.
The grid of $\lambda$ consists of $16$ values evenly spaced in log-space between $10^{-4}$ and $10$, as well as $0$. 

\section{Additional Experiment}
\label{app:sec:extra}
We present the results of another experiment, where all methods, including Bloop, gave similar Pareto fronts.
Here, we aim to perform classification on both the Imagenet and the CIFAR-10 datasets.
The network is a ResNet50 with with two separate classification heads.
This problem sits in the middle ground between the multi-task learning and the joint dataset training problem that we describe in \cref{sec:expe}: we have two separate datasets for the two distinct tasks.
Similar to before, the main loss is the training loss on the larger dataset, i.e., Imagenet, and the auxiliary loss is the training loss on the smaller dataset, i.e. Cifar10.
We choose $\lambda$ to be equally split on log scale from $10^{-3}$ to $10$.
The remaining configurations follow the experiment of Imagenet training with L2 regularization, except that we also scale the learning rate by $1/(1+\lambda)$ to avoid instability as in the multi-task experiment.

\begin{figure}
    \centering
    \includegraphics[width=0.24\textwidth]{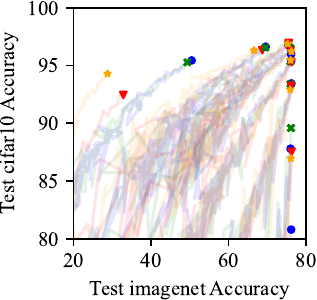}
    \includegraphics[width=0.24\textwidth]{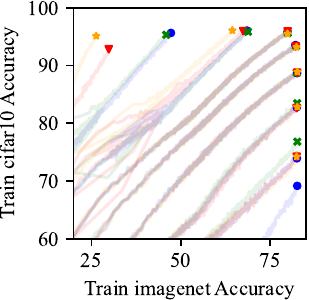}
    \includegraphics[width=0.24\textwidth]{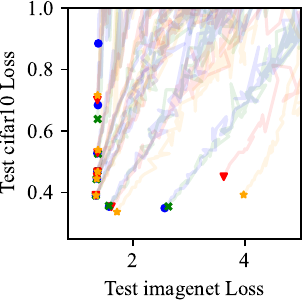}
    \includegraphics[width=0.24\textwidth]{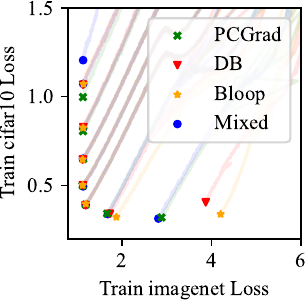}
    \caption{Results on the Imagenet / Cifar10 experiment. All algorithms perform generally similarly except for very high values of $\lambda$, which leads to worse performance for all algorithms.}
    \label{app:fig:cifarimagenet}
\end{figure}

The results are shown in \autoref{app:fig:cifarimagenet}. Unlike the experiments of \cref{sec:expe}, there is little trade-off between the two tasks. We can increase accuracy on CIFAR-10 without sacrificing performance on Imagenet.
For this reason, there are only very few points at the Pareto front and all methods perform similarly at these points.
We posit that here, the two losses are not conflicting enough to see the gradient surgery methods have an edge. 

\end{document}